%% file: main.tex
\documentclass[10pt,twocolumn,letterpaper]{article}
\pdfoutput=1 
\usepackage[pagenumbers]{iccv} 


\usepackage{bbding}
\usepackage{multirow}
\usepackage{multicol}
\usepackage{paralist}
\usepackage{xcolor}
\usepackage{colortbl}
\usepackage{relsize}

\definecolor{baselinecolor}{gray}{.9}
\newcommand{\baseline}[1]{\cellcolor{baselinecolor}{#1}}

\definecolor{iccvblue}{rgb}{0.21,0.49,0.74}
\usepackage[pagebackref,breaklinks,colorlinks,allcolors=iccvblue]{hyperref}

\def\paperID{8611} 
\def\confName{ICCV}
\def\confYear{2025}

\title{VLForgery Face Triad: Detection, Localization and Attribution via Multimodal Large Language Models}

\author{
Xinan He$^{1}$, Yue Zhou$^{2}$\thanks{The author contributed equally to this work.}, Bing Fan$^{3}$, Bin Li$^{2}$, Guopu Zhu$^{4}$, Feng Ding$^{1}$\thanks{Corresponding author}\\
$^{1}$Nanchang University \tt \small shahur@email.ncu.edu.cn; fengding@ncu.edu.cn\\
$^{2}$Shenzhen University \tt \small 2450042008@email.szu.edu.cn; \tt \small libin@szu.edu.cn\\
$^{3}$University of North Texas \tt \small bingfan@my.unt.edu\\
$^{4}$Harbin Institute of Technology \tt \small guopu.zhu@hit.edu.cn\\
}

\begin{document}
\maketitle
\input{sec/0_abstract}    
\input{sec/1_intro}
\input{sec/2_relat}

\input{sec/3_VLFCon}
\input{sec/4_VLFBench}
{
    \small
    \bibliographystyle{ieeenat_fullname}
    \bibliography{main}
}


\end{document}

%% file: sec/0_abstract.tex
\begin{abstract}
Faces synthesized by diffusion models (DMs) with high-quality and controllable attributes pose a significant challenge for Deepfake detection. Most state-of-the-art detectors only yield a binary decision, incapable of forgery localization, attribution of forgery methods, and providing analysis on the cause of forgeries. In this work, we integrate Multimodal Large Language Models (MLLMs) within DM-based face forensics, and propose a fine-grained analysis triad framework called \textbf{VLForgery},
that can 1) \textcolor{blue}{predict falsified facial images};
2) \textcolor{blue}{locate the falsified face regions} subjected to partial synthesis; and 3) \textcolor{blue}{attribute the synthesis} with specific generators. To achieve the above goals, we introduce \textbf{VLF} (Visual Language Forensics), a novel and diverse synthesis face dataset designed to facilitate rich interactions between `\textbf{V}isual' and `\textbf{L}anguage' modalities in MLLMs.
Additionally, we propose an extrinsic knowledge-guided description method, termed \textbf{EkCot}, which leverages knowledge from the image generation pipeline to enable MLLMs to quickly capture image content. Furthermore, we introduce a low-level vision comparison pipeline designed to identify differential features between real and fake that MLLMs can inherently understand. These features are then incorporated into EkCot, enhancing its ability to analyze forgeries in a structured manner, following the sequence of detection, localization, and attribution.
Extensive experiments demonstrate that VLForgery outperforms other state-of-the-art forensic approaches in detection accuracy, with additional potential for falsified region localization and attribution analysis. 
\end{abstract}


%% file: sec/1_intro.tex
\vspace{-2mm}
\section{Introduction}
\vspace{-2mm}
\label{sec:intro}

\begin{figure}[t]
  \centering
  \includegraphics[width=0.98 \linewidth]{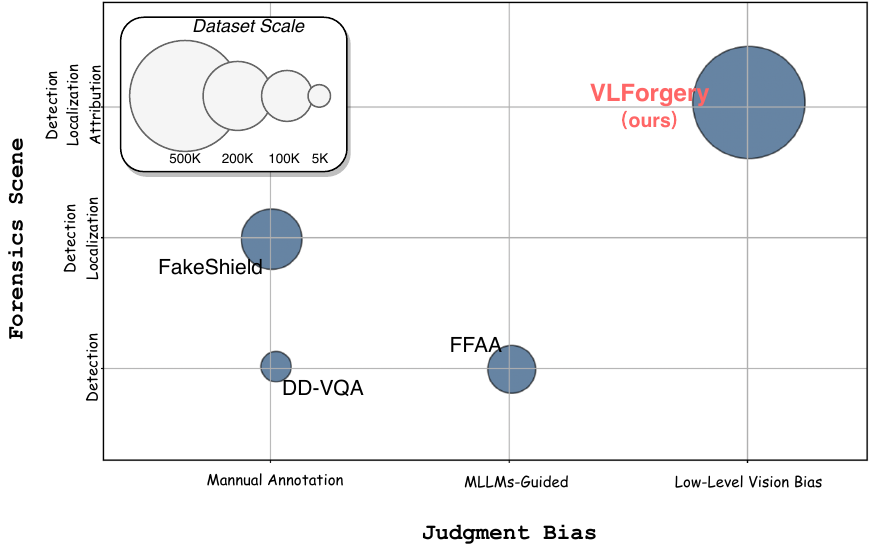}
  \vspace{-3mm}
  \caption{Comparison of VLForgery with other MLLMs forensics frameworks.}
  \vspace{-5mm}
  \label{fig:introduction}
\end{figure}

AI-generated faces have enriched human life by creating realistic virtual characters that enhance our experiences in various creative industries \cite{feng2024deepfakesurvey}. However, the domain of AI-generated face forensics is encountering unprecedented challenges due to the rapid advancement of generative models \cite{fan2024synthesizing}. In particular, diffusion models \cite{sohl2015deep,rombach2022high} generate synthetic faces with substantially higher fidelity than earlier technologies such as variational autoencoders (VAEs) \cite{vahdat2020nvae, ding2022securing}, rendering it increasingly challenging for detectors to identify subtle forgery artifacts. This development has led to a significant surge in the proliferation of forgeries, which present grave threats to personal privacy and security \cite{ding2021anti, ding2024disrupting}.

Several pioneering studies \cite{chen2024drct, khan2024clipping, wang2023dire} have investigated diffusion model (DM)-generated image forensics and demonstrated the effectiveness of their proposed models.
However, a notable limitation of these approaches is their lack of clear analysis for the underlying causes of the forgeries. 
Specifically,\textit{ most existing studies focus solely on detection, neglecting the localization of forgeries or attributing specific generation methods to them.}

Recently, multimodal large language models (MLLMs) have emerged as powerful tools for complex scene characterization.
By leveraging extensive pre-training on both images and text, MLLMs can interpret visual tasks through natural language, providing a more nuanced understanding of forensic analysis \cite{zhang2024common, huang2024ffaa}. However, MLLM-based models face \textit{\textbf{two}} primary limitations. \textbf{First}, forensic analysis research with MLLMs has largely focused on traditional face-swapping Deepfakes \cite{Rossler_Cozzolino_Verdoliva_Riess_Thies_Niessner_2019, li2020celeb}, 
with limited analysis of DM-based synthetic faces. While recent work \cite{zhang2024common, huang2024ffaa} has shown that automated descriptions generated by MLLMs (\eg, ChatGPT-4) or manual annotations guiding MLLM training, effectively capture artifacts in traditional face-swapping Deepfakes, these approaches still face challenges in generating credible forgery descriptions, particularly for high-fidelity DM-based images where subtle artifacts often evade detection. The primary issue lies in their reliance on subjective human-defined judgment biases to guide the generation of descriptions. Such descriptions may not align with the forgery knowledge that the models inherently understand. Additionally, MLLMs are typically trained for semantic-level visual alignment, lacking fine-grained forensic perception capabilities, which can lead to hallucinations in the generated descriptions. \textbf{Second}, none of the existing MLLM-based methods attempt fine-grained DeepFake forensics that simultaneously addresses \textit{Detection}, \textit{Localization}, and \textit{Attribution}. 

To address these challenges, we propose a triad framework-encompassing detection, localization, and attribution-designed for AI-generated face forensics, referred to as \textbf{VLForgery}. 
In the proposed framework, to address the lack of data specifically tailored for DM-based partial synthesis face,
we create diverse prompt repositories and templates to build a new multimodal \underline{V}isual \underline{L}anguage \underline{F}orensic (\textbf{VLF}) dataset, suitable for the triad tasks of detection, localization, and attribution.
Second, to enhance the reliability of MLLMs-generated descriptions, we introduce a low-level vision comparison pipeline that identifies low-level vision discrepancies between real and fake sample sets, guiding MLLMs to generate high-confidence descriptions. Building on this, we develop a description generation module, where we design an \underline{E}xtrinsic \underline{k}nowledge-guided \underline{C}hain-\underline{o}f-\underline{t}hought (\textbf{EkCot}) method. This method also integrates additional knowledge from the generation pipeline, enabling the model to quickly grasp image content.
Third, we develop a unified MLLMs fine-tuning and inference module, where we aggregate the generated images and corresponding descriptions to fine-tune MLLMs.

Specifically, VLForgery incorporates external information relevant to DM-based image generation with the low-level vision discrepancies between real and fake to structure a systematic analytical approach, allowing MLLMs to evaluate the image through the following steps: 1) \textit{Detection}:  identify whether the image is real or AI-generated; 2) \textit{Localization}:  For synthetic images, determine if the forgery is partial (only specific regions are altered) or full (the entire image is synthetic). For partial forgeries, locate the altered regions;  3) \textit{Attribution}:  Determine the likely method or model used to generate the forgery.

Our contribution can be summarized as follows:
\begin{compactitem}
\item We investigate the potential of MLLMs in tackling AI-generated face forensics challenges by proposing the framework VLForgery. Additionally, we designed an extrinsic knowledge-guided chain-of-thought method, termed EkCot, that can assist the MLLMs in achieving fine-grained forensics performance.
\item We introduce the VLF dataset, a new multimodal 
 Visual Language Forensic dataset generated by diffusion models, designed to address the needs of fine-grained forensics tasks.
\item We present a comprehensive evaluation experiment tailored to assess VLForgery on fine-grained forensic tasks. This evaluation encompasses three task scenarios, involving 5 traditional deepfake types and 9 DM-based face types.
\end{compactitem}

%% file: sec/2_relat.tex
\begin{figure*}[t]
  \centering
  \includegraphics[width=\textwidth]{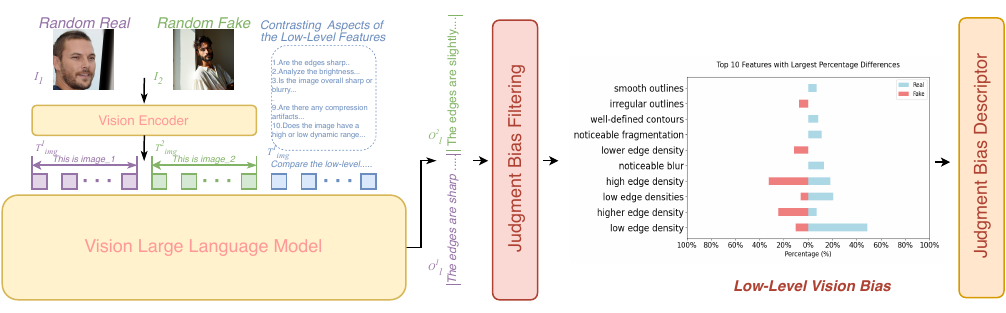}
  \vspace{-6mm}
  \caption{The comparison pipeline of low-level vision. By extracting the most distinctive low-level vision descriptors that differentiate the real and fake sample sets, we establish judgment biases for the MLLMs.}
  \label{fig:LowLevelCom}
  \vspace{-4mm}
\end{figure*}
\vspace{-1mm}
\section{Related Work}

\label{sec:related work}

\vspace{-1mm}
\subsection{DM-Generated Images }
\vspace{-1mm}
Recent advancements have centered on diffusion models \cite{sohl2015deep}, which have demonstrated exceptional performance in generating high-fidelity face images with high quality. Many studies have also incorporated attention mechanisms to
enhance the controllability of generated content \cite{ku2023imagenhub, lin2024detecting} such as text-guided \cite{rombach2022high,mokady2023null,wu2023latent,parmar2023zero,zhang2024magicbrush,brooks2023instructpix2pix} and subject-guided \cite{ruiz2023dreambooth,li2023dreamedit,zhang2023adding,kawar2023imagic}.
These achievements have led to the proposal of many DM-based datasets, such as object-oriented image datasets \cite{zhu2024genimage, corvi2023detection, stockl2023evaluating} and face-oriented image datasets \cite{cheng2024diffusion, lin2024ai}. However, most DM-based partially synthesized face datasets suffer from poor quality and lack sufficient ground truth to support more fine-grained forensics tasks, such as forgery localization and attribution.

\vspace{-1mm}
\subsection{Fine-grained Forensics of DM-based Images}
\vspace{-1mm}
\underline{For Detecting tasks}, traditional works usually attempt to find the traces left by diffusion models to expose synthesized images. For example, 
Chen \emph{et al.} \cite{chen2024drct} proposed DRCT, a universal framework to enhance the generalizability of the existing detectors for detecting DM-based images. 
Recognizing the robust semantic comprehension capabilities of pre-trained vision-language models (\eg, CLIP \cite{radford2021learning}), recent studies \cite{khan2024clipping,ojha2023towards} have demonstrated their promising results in detecting DM-generated images. \underline{For Locating tasks}, a primary goal is for the detector to provide predictions with a localization map indicating which regions have been manipulated or to identify forgery artifacts within DM-based images. Guo et al. \cite{guo2023hierarchical} proposed a hierarchical fine-grained Image Forgery Detection and Localization (IFDL) framework consisting of three components: a multi-branch feature extractor, localization, and classification modules. Zhang \emph{et al.} \cite{zhang2023perceptual} centers on detecting and segmenting artifact areas that are only perceptible to human observers, rather than identifying the full manipulation region.
\underline{For Attributing tasks}, the objective is to recognize the specific diffusion model that generates the images. Guarnera et al. \cite{guarnera2024level} concentrate on attributing DM-generated images through a multi-level hierarchical approach. However, the aforementioned studies address the tasks of detection, localization, and attribute individually, and there is a notable absence of research that integrates these tasks into a unified framework.

\subsection{Multimodal Large Language Models in Forensics}
\vspace{-1mm}
Recently, some studies have commenced utilizing MLLMs to investigate analytical ability within the field of forensics. Jia \emph{et al.} \cite{jia2024can} pioneered the exploration of the forensic capabilities of prompt engineering using ChatGPT. Zhang \emph{et al.} \cite{zhang2024common} proposed incorporating common-sense reasoning to enhance traditional deepfake detection by manually labeling the manipulated regions and extending it to develop a Visual Question Answering (VQA) framework.
Huang \emph{et al.} \cite{huang2024ffaa} recently proposed an automatic method utilizing GPT-4o for annotations, thereby replacing human annotators.
These studies primarily focus on traditional Deepfakes and are limited to single-forensic scenarios, lacking the capability to locate or attribute forgery areas in synthesized images. In the case of forgery artifacts from high-fidelity faces generated by diffusion models, generating credible forgery analyses may pose greater challenges. As illustrated in Fig.~\ref{fig:introduction}, ours, compared to other MLLM-based forensic frameworks, can guide MLLMs to generate highly credible fine-grained forgery descriptions and is applicable to a broader range of forgery types and more complex forensic scenarios.


%% file: sec/3_VLFCon.tex
\begin{figure*}[t]
  \centering
  \includegraphics[width=\textwidth]{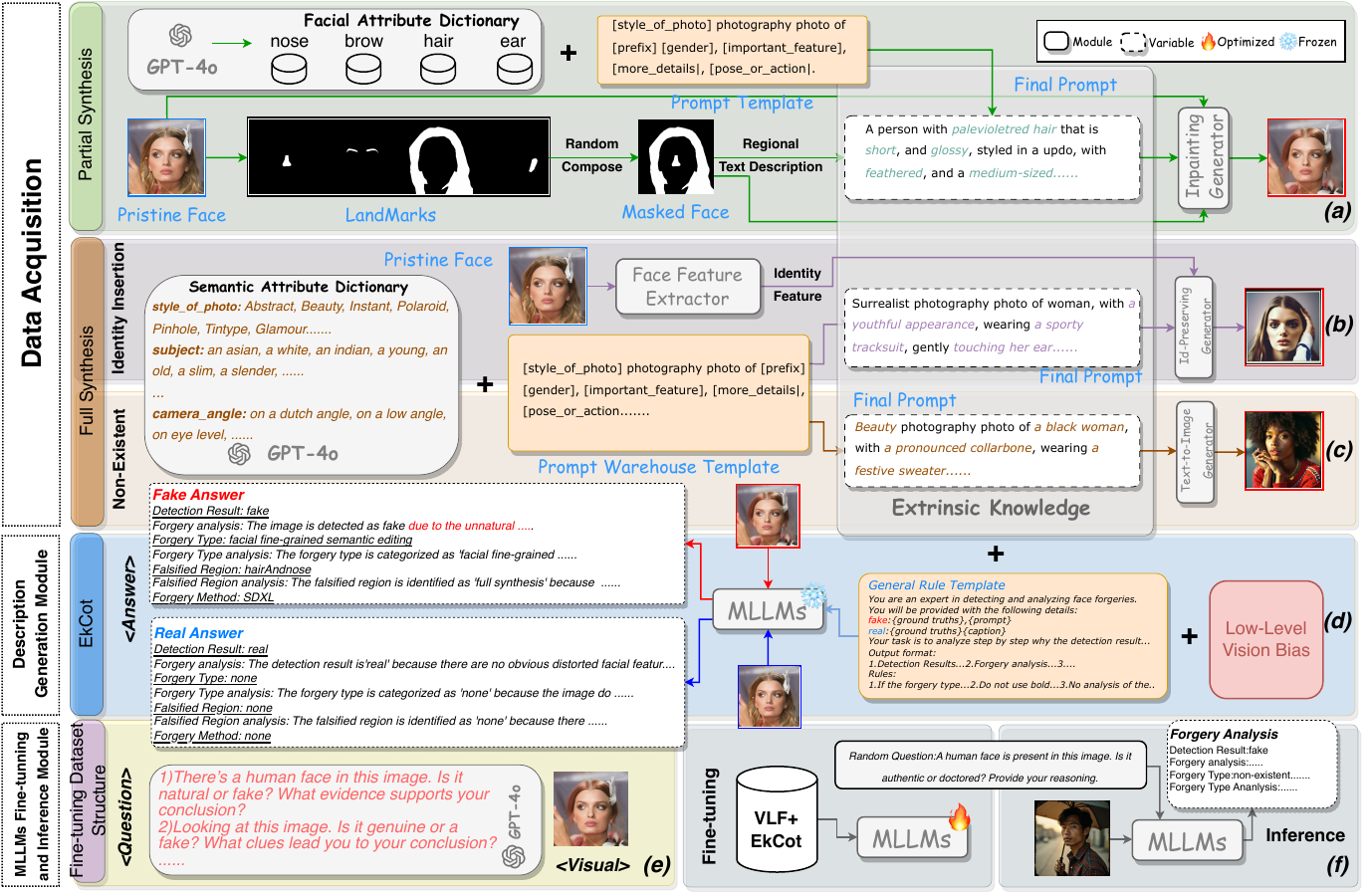}
  \vspace{-6mm}
  \caption{Construction of three modules for VLForgery framework: Data Acquisition, Description Generation Module, and MLLMs Fine-tuning and Inference Module.}
  \label{fig:Fig3}
  \vspace{-6mm}
\end{figure*}

\vspace{-1mm}
\section{VLForgery Framework}
\label{sec:VLF Construction}
\vspace{-1mm}
With the development of generative technologies, methods like diffusion models have emerged. Compared with traditional generative AI techniques (\eg, VAE), the diffusion models enable partial synthesis guided by masks and text descriptions, as well as full synthesis through identity insertion or text descriptions. Therefore, our VLF dataset encompasses both partial and full synthesis types. 
In the following sections, we will demonstrate how to acquire their corresponding prompts and/or images. Furthermore, we will elaborate on the specific details of our proposed Description Generation Module and MLLMs Fine-tunning and Inference Module in VLForgery. 
\vspace{-1mm}
\subsection{Data Acquisition}
\vspace{-1mm}
This section primarily outlines the pipeline details for collecting data.
We synthesized all DM-generated data according to the requirements of the fine-grained forensics tasks. The details of the generation, involving partial and full synthesis, will be described in the following sections.
\vspace{-1mm}
\subsubsection{Partial Synthesis}
\vspace{-1mm}
In this part, we define the pipeline for partial synthesis. An exemplar pipeline is illustrated in {Fig.~\ref{fig:Fig3}(a)}.

To ensure the high quality of the partially synthesized images, we utilized the well-established facial dataset CelebAMask-HQ \cite{CelebAMask-HQ} as a data source since it already contains landmarks for various face attributes such as nose, brow, hair, ears, eyes, and teeth. Then, we use GPT-4o to generate each facial attribute dictionary and prompt template. For each image, we randomly combine masks to generate masked faces. Based on the generated masked face, we choose and combine templates for the corresponding attributes, then randomly select descriptive keywords from the relevant attribute dictionary to populate the templates, thereby generating the final prompt for image partial synthesis. During the image generation phase, we input the generated masked face, corresponding final prompt, and pristine face into the Inpainting Generator to produce an inpainted face. 

\vspace{-1mm}
\subsubsection{Full Synthesis}
\vspace{-1mm}

We introduce two types of fully synthesized forgery: generating identity insertion (Fig.~\ref{fig:Fig3}(b)) and non-existent faces (Fig.~\ref{fig:Fig3}(c)). Both generation approaches utilize the same prompt template with slight modifications. In the face identity insertion pipeline, the objective is to preserve the original facial identity while situating it within a new scene. This pipeline initially uses a Face Feature Extractor to capture identity-specific features. In the generation of non-existent faces, the diffusion model creates images of completely new faces that are unrelated to any existing individual based on a given prompt.

\vspace{-1mm}
\subsection{Description Generation Module}
\vspace{-1mm}
This module leverages the extrinsic information required throughout the low-level vision comparison pipeline and the data acquisition processes to guide the construction of the forgery analysis description for each type of forgery sample in the VLF dataset. In this section, we introduce the workflows of the low-level vision comparison pipeline and the construction of EKCot separately. \textit{We provide a detailed low-level vision analysis along with all the details and data you may need.}
\subsubsection{Low-level Feature Comparison Pipeline}
\vspace{-1mm}
\smallskip
\noindent
\underline{Motivation:} In previous studies\cite{xu2024fakeshield, huang2024ffaa}, researchers attempted to generate explainable descriptions for Deepfakes using Vision Large Language Models. However, hallucinations are commonly observed in VQA tasks. Forensic analysis places more emphasis on the perception of image fine-grained details compared with general VQA tasks. Therefore, in the absence of ground-truth labels to supervise these explainable descriptions, \textbf{how can we determine their correctness?}

To address this challenge, we aim to seek forensic features that the model itself can understand to discriminate the authenticity of an image, thereby mitigating the model's struggles with explainable description generation. Therefore, we propose a low-level vision comparison pipeline that leverages the visual feature understanding capabilities of vision-large-language models (VLLMs) to identify low-level vision discrepancies between real and fake sample sets, as illustrated in Fig.~\ref{fig:LowLevelCom}. First, we randomly select one image from the real sample set and another from the fake sample set, as $I_1$ and $I_2$. Notably, the VLLMs is unaware of which image is real and which is fake.
Furthermore, to comprehensively verify the low-level visual discrepancies between the real and fake sample sets, we selected ten distinct aspects and designed corresponding descriptive templates for each, as $T_{tem}$.

Initially, $I_1$ and $I_2$ are processed by the vision encoder $E_v$, which generates the corresponding token representations $T_{img}$. These tokens, along with the comparison template tokens $T_{tem}$, are then fed into the $VLLM()$ for generating low-level vision discrepancies text $O_l$. It is formulated as:
\vspace{-1mm}
\begin{subequations}
    \begin{align}
    &T^1_{img}, T^2_{img} = E_v(I_1, I_2),
    \\
    &O^1_l, O^2_l = VLLM(T^1_{img}, T^2_{img}, T_{tem}).
    \end{align}
\end{subequations}
\vspace{-1mm}
\smallskip
\noindent
\underline{Judgment Bias filtering:} Given $O^1_l$,$O^2_l$, we first perform tokenization and part-of-speech tagging to extract compound noun phrases. Next, we compute the proportion of each extracted compound noun phrase across all low-level vision discrepancy texts. Finally, we identify the top ten phrases with the largest proportion difference between the real and fake sample sets, selecting them as the judgment biases for the MLLMs. These phrases are then used to generate the final usable judgment bias descriptors through a predefined template.

Furthermore, to avoid potential model interpretation biases, we selected the Llama-3.2-11B-Vision\cite{Llama-3.2-11B-Vision} model as the VLLM for the low-level vision comparison pipeline, ensuring consistency with the models used in subsequent stages.

\vspace{-1mm}
\subsubsection{Construction of EkCot}
\vspace{-1mm}
As shown in Fig.~\ref{fig:Fig3}(d), we utilized MLLMs as the description generation model to construct a chain of thought for forgery analysis, termed EkCot. Initially, we propose a General Rule template designed to integrate the extrinsic information required for generating desired data. The extrinsic information includes the prompts used within the pipeline, the corresponding ground truth for the generated image, and the low-level vision judgment bias, which are subsequently combined to serve as prompt inputs for MLLMs. The ground truth encompasses several categories, including real/fake, forgery type, forgery region, and forgery method. For pristine images, the forgery type, forgery region and forgery method values are set to 'none' by default. Additionally, image original features are provided as visual inputs to the MLLMs. Furthermore, we utilize the Llama-3.2-11B-Vision\cite{Llama-3.2-11B-Vision} as the description generation model.
\vspace{-1mm}
\subsection{MLLMs Fine-tuning and Inference Module}
\vspace{-1mm}

In this module, we designed the final fine-tuning data structure and the details of MLLMs fine-tuning.
\smallskip
\noindent
\textbf{Final Fine-tuning Data Structure.} The final fine-tuning data is composed in the form of triplets: $<$visual, question, answer$>$. 
\smallskip
\noindent
\underline{For question's format}, we use ChatGPT-4o to generate a range of question formats, as shown in Fig.~\ref{fig:Fig3}(e), for example: `Is this image real or fake? Can you provide the reasoning behind your judgment?' For each triplet, the question is presented in a randomly generated format. Additionally, each image is linked to its corresponding specific description (i.e., the answer).
\smallskip
\noindent
\underline{For the answer's format}, we designed it to address three primary forensic tasks: detection, localization, and attribution. The answer format follows this pattern: \textit{`Detection Result: real/fake'} + $A_{result}$ + \textit{`Forgery Type'} + \textit{`Falsified Region'} + \textit{`Forgery Method'}, where $A_{result}$ denotes the analysis of detection results. 
\smallskip
\noindent
\underline{For analysis steps}, our objective is to guide the model to adopt a chain of thought, performing each step sequentially as follows: 1)Real or Fake Judgment: The authenticity of the image is first assessed. If an image is identified as a forgery, the type of forgery is then classified into one of three categories: partial synthesis, identity insertion, and non-existent faces. The reasons for the detection result are further analyzed. 2) Localization of Falsified Region: If the forgery involves partial synthesis, the specific regions of manipulation (\textit{e.g.}, nose, brow, hair, ear) are identified. In cases involving multiple edits (\textit{e.g.}, hairAndnose), each forged region is localized. 3) Attribution of Synthesis Face Generator: The final step determines the generator of the synthesis face.
\smallskip
\noindent
\textbf{Fine-tunning and Inference Process.}
We fine-tune the MLLMs on the final fine-tuning data through a structured two-step process (Fig.~\ref{fig:Fig3}(f)). Inspired by LLaVA \cite{liu2024visual}, we adopt their published Llava-1.5-7B model and training architecture. First, we fine-tune a projector to align the facial visual features extracted from the frozen CLIP visual encoder with the corresponding question text features. This alignment links the synthetic face’s forgery artifacts with corresponding ground truth and detailed forgery descriptions. Second, we employ Low-Rank Adaptation (LoRA) \cite{hu2021lora} (rank=128, alpha=256) to selectively update the model by adjusting only the LoRA parameters, effectively fine-tuning the pre-trained language model.

During inference, the input is structured as a tuple $<$image, question$>$. For the questions, we continue to use ChatGPT-4 to generate diverse question formats, ensuring input variety. Finally, the model's response is structured, similar to the final fine-tuning data‘s answer format.

This approach ensures the model conducts forensic analysis on input faces in a structured, stepwise manner according to a pre-defined reasoning framework, thereby improving both its fine-grained forgery analysis and forensic accuracy.

%% file: sec/4_VLFBench.tex
\begin{table}[th]  
\centering
\scalebox{0.65}{
    \begin{tabular}{cccc}
    \toprule[1.3pt]
     \multirow{2}[1]{*}{Type}& \multirow{2}[1]{*}{Source}&\multicolumn{2}{c}{Samples}\\ \cmidrule(r){3-4}
     &&train&test\\ \hline
    
    \multirow{3}{*}{Pristine}
    &FFHQ&42k&14k\\
    &FF++ (original)&12k&4.5k\\
    &CelebA-HQ&18k&6k\\ \hline
    \multirow{3}{*}{DM-based Partial Synthesis}
    &SDXL&76k&19k\\
    &SD2&76k&19k\\
    &Kandinsky2.2&76k&19k\\ \hline
    \multirow{6}{*}{DM-based Full Synthesis}
    &SDXL&24k&6k\\
    &SD2&24k&6k\\
    &SD3&24k&6k\\
    &Kandinsky2.2&24k&6k\\
    &Flux&24k&6k\\ 
    &InstantID&8.1k&2.5k \\ 
    \bottomrule[1.3pt]
    \end{tabular}
}
\vspace{-3mm}
\caption{VLF dataset.}
\label{tab:VLFSources}
\vspace{-6mm}
\end{table}

\section{Experiment}

\begin{figure*}[htbp]
  \centering
  \includegraphics[width=\textwidth]{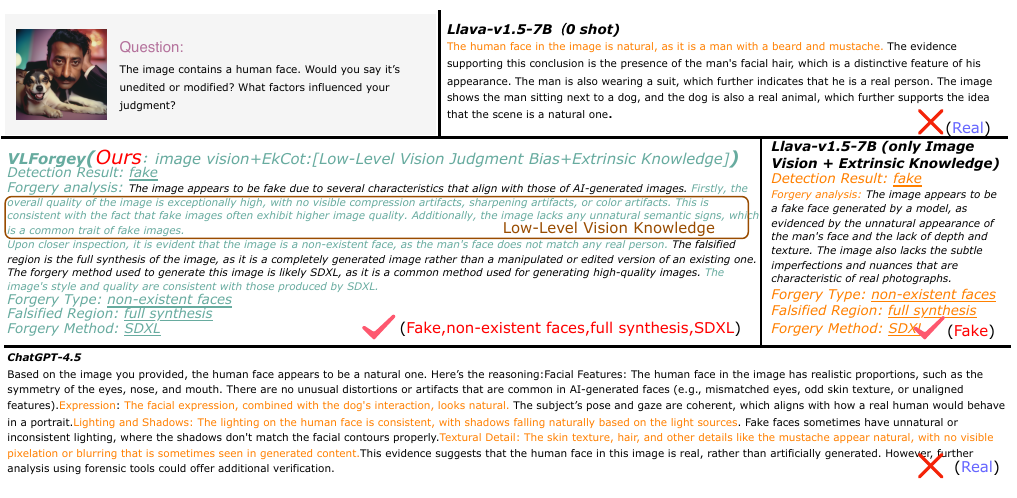}
  \vspace{-8mm}
  \caption{Qualitative results. The upper left presents a synthesis face from full synthesis. Ours is presented in contrast to the responses of Llava-v1.5-7B (0 shot), Llava-v1.5B-7B (conditional fine-tune), and Chatgpt-4.5, respectively. The lower-right bracket provides the result annotations.}
  \label{fig:comparison with advanced LLMs}
  \vspace{-5mm}
\end{figure*}






\subsection{Experimental Settings}
\paragraph{Datasets.}
All facial samples in experiments were sourced from the VLF. The VLF dataset consists of 12 subsets, with each subset divided into training and test sets in a 4:1 ratio, as illustrated in Table~\ref{tab:VLFSources}.


\paragraph{Compared Baselines.}
We selected four types of models for evaluation: 1) Naive convolutional neural networks (CNNs): Xception \cite{Chollet2017},  Resnet-50 \cite{He_Zhang_Ren_Sun_2016}. 2) Typical frequency detectors: SPSL \cite{liu2021spatial}, F3Net\cite{qian2020thinking}, SRM\cite{luo2021generalizing}, NPR\cite{tan2024rethinking}. 3) Typical Spatial detectors: GramNet\cite{liu2020global}, SAFE\cite{li2024improving}, CNNspot\cite{wang2020cnn}, DRCT \cite{chen2024drct}.  4) VLLM-based method:CLIPping \cite{khan2024clipping}, Llava-1.5-7B\cite{liu2024visual}, Qwen2.5-VL-7B\cite{yang2024qwen2}, Llama3.2-11B-Vision\cite{Llama-3.2-11B-Vision}.

\paragraph{Implementation Details.} All experiments are based on the PyTorch and trained with 8 NVIDIA RTX L40. For training, we utilized the Adam optimizer with a learning rate of 2e-5 and a batch size of 128 for 3 epochs.

\begin{table}[ht!]
\centering
\scalebox{0.60}{
\begin{tabular}{c|cccccc|c}
\toprule[1.3pt]
\multirow{3}{*}{Train} &
\multicolumn{6}{c|}{Full Synthesis(\%)\textuparrow}&\multirow{3}{*}{Avg(\%)\textuparrow} \\ \cline{2-7}

& \multicolumn{5}{c|}{Non-existent} & Id-insertion& \\ \cline{2-7}

&SDXL&SD2&SD3&Kandinsky2.2&\multicolumn{1}{c|}{Flux}&InstantID& \\ \hline

SDXL&\baseline{99.35}&98.47&96.02&98.78&98.7&87.38&96.45\\ 
SD2&99.47&\baseline{\textbf{99.97}}&98.38&99.40&98.92&80.67&96.13\\ 
SD3&99.65&97.23&\baseline{\textbf{99.98}}&97.95&99.70&21.93&86.07\\ 
Kandinsky2.2&\textbf{99.95}&99.37&97.95&\baseline{\textbf{99.95}}&99.67&\textbf{93.75}&\textbf{98.44}\\ 
Flux&98.01&95.85&98.10&98.25&\baseline{\textbf{100.00}}&48.10&74.89\\
InstantID&90.38&87.05&60.83&87.40&84.97&\baseline{92.97}&83.94\\ \bottomrule[1.3pt]

\end{tabular}
}
\vspace{-3mm}
\caption{Evaluation on Full Synthesis type. Results on different training and testing subsets using VLForgery. Accuracy is used for evaluation.}
\label{tab:task1_llava_fullsynthesis}
\vspace{-6mm}
\end{table}

\begin{table}[ht!]
\centering
\scalebox{0.60}{
\begin{tabular}{c|ccc|c}
\toprule[1.3pt]
\multirow{2}{*}{Train} &
\multicolumn{3}{c|}{Partial Synthesis(\%)\textuparrow}&\multirow{2}{*}{Avg(\%)\textuparrow} \\ \cline{2-4}
&SDXL&SD2&\multicolumn{1}{c|}{Kandinsky2.2}&\\ \hline

SDXL&\baseline{\textbf{80.35}}&\textbf{68.78}&\textbf{98.08}&\textbf{82.40}\\ 
SD2&54.53&\baseline{61.77}&93.01&69.77\\ 
Kandinsky2.2&42.09&34.20&\baseline{95.93}&57.41\\ \bottomrule[1.3pt]

\end{tabular}
}
\vspace{-3mm}
\caption{Evaluation on Partial Synthesis type. Results on different training and testing subsets using VLForgery. Accuracy is used for evaluation.}
\label{tab:task1_llava_partialsynthesis}
\vspace{-6mm}
\end{table}

\subsection{Evaluation of Multi-task Forensics}

\subsubsection{Task1: Detecting the Authenticity of Faces}

\begin{table*}[ht!]
\centering
\scalebox{0.73}{
\begin{tabular}{c|c|cc|ccccccccccccc}
\toprule[1.3pt]
\multirow{2}{*}{Type}&\multirow{2}{*}{Method} &
\multicolumn{2}{c|}{Train--SDXL(\%)\textuparrow}& \multicolumn{2}{c}{SD2(\%)\textuparrow}&\multicolumn{2}{c}{Kandinsky2.2(\%)\textuparrow}&\multirow{2}{*}{SD3(\%)\textuparrow}&\multirow{2}{*}{Flux(\%)\textuparrow}&\multirow{2}{*}{InstantID(\%)\textuparrow}&\multirow{2}{*}{Avg(\%)\textuparrow} \\ \cmidrule(r){3-4} \cmidrule(r){5-6} \cmidrule(r){7-8} 

&&PS&FS&PS&FS&PS&FS&&& \\  \hline

\multirow{2}{*}{Naive}&Resnet-50\cite{He_Zhang_Ren_Sun_2016}&35.09&99.73&11.87&95.67&58.40&97.37&90.58&94.03&94.58&75.26\\ 
&Xception\cite{Chollet2017}&68.19&99.98&40.10&97.58&96.89&99.68&97.72&99.03&99.57&88.75\\

\hline
\multirow{4}{*}{Frequency}&SPSL\cite{liu2021spatial}&31.62&99.98&4.79&96.92&51.42&99.83&96.92&99.28&95.09&75.09\\
&F3Net\cite{qian2020thinking}&74.09&99.91&46.97&99.45&81.65&99.90&99.33&99.98&99.84&89.01\\
&SRM\cite{luo2021generalizing}&59.21&99.95&12.30&98.65&98.34&99.57&98.43&98.17&97.06&84.63\\
&NPR\cite{tan2024rethinking}&42.67&99.98&10.22&98.98&70.51&99.57&98.25&99.75&99.37&79.92\\

\hline
\multirow{4}{*}{Spatial}&GramNet\cite{liu2020global}&34.17&99.98&14.28&96.93&48.37&99.65&93.43&99.27&88.45&74.94\\
&SAFE\cite{li2024improving}&56.47&82.08&54.08&81.55&56.32&78.98&88.43&57.05&71.43&61.66\\
&CNNspot\cite{wang2020cnn}&26.30&97.93&9.04&73.00&26.97&74.12&39.95&70.00&73.60&54.55\\ 
&DRCT$^*$\cite{chen2024drct}&57.65&99.41&58.52&97.71&99.33&99.92&65.97&59.06&98.92&81.83\\

\hline
\multirow{6}{*}{VLLM-Based}&CLIPping \textsmaller{\emph{Adapter}}\cite{khan2024clipping}&36.29&99.91&32.13&94.37&75.04&98.38&98.58&97.95&97.05&81.03\\ 
&CLIPping \textsmaller{\emph{Linear Probing}}\cite{khan2024clipping}&61.79&99.98&54.33&97.18&87.33&99.65&97.88&99.03&99.80&88.55 \\
&Llava-1.5-7B\cite{liu2024visual}&69.92&99.92&51.07&99.97&99.56&99.74&97.66&99.12&99.69&87.07 \\ 
&Qwen2.5VL-7B$^*$\cite{yang2024qwen2}&4.77&4.32&3.78&21.88&31.92&6.55&3.82&3.00&46.13&14.02\\
&Llama-3.2-11B-Vision$^*$\cite{Llama-3.2-11B-Vision}&53.46&64.97&52.13&68.95&58.42&69.63&47.80&63.05&58.49&59.66\\
\cline{2-12}
&\baseline{\begin{tabular}[c]{@{}c@{}}VLForgery\\ (ours)\end{tabular}}&\baseline{78.62}&\baseline{99.98}&\baseline{66.32}&\baseline{99.97}&\baseline{99.96}&\baseline{99.97}&\baseline{99.98}&\baseline{99.97}&\baseline{99.57}&\baseline{\textbf{93.82}}\\ 
\bottomrule[1.3pt]

\end{tabular}
}
\vspace{-3mm}
\caption{Cross-generator dataset evaluation on ACC metric. All methods are trained on SDXL and evaluated on other subsets. $*$ indicates the use of trained models provided by the authors. Note that `PS' means Partial Synthesis, and `FS' means Full Synthesis.
}
\label{tab:task1_llava_anydetector}
\vspace{-4mm}
\end{table*}

\begin{table*}[ht!]
\centering
\scalebox{0.60}{
\begin{tabular}{c|cccc|cccc|cccc|cccc}
\toprule[1.3pt]
\multirow{2}{*}{Train} &
\multicolumn{3}{c}{hair(\%)\textuparrow}&\multirow{2}{*}{Avg(\%)\textuparrow}& \multicolumn{3}{c}{brows(\%)\textuparrow}&\multirow{2}{*}{Avg(\%)\textuparrow}&\multicolumn{3}{c}{ears(\%)\textuparrow}&\multirow{2}{*}{Avg(\%)\textuparrow}& \multicolumn{3}{c}{nose(\%)\textuparrow} & \multirow{2}{*}{Avg(\%)\textuparrow}\\ \cmidrule(r){2-4} \cmidrule(r){6-8} \cmidrule(r){10-12} \cmidrule(r){14-16}

&SDXL&SD2&\multicolumn{1}{c}{Kandinsky2.2}&& SDXL&SD2&Kandinsky2.2&&SDXL&SD2&\multicolumn{1}{c}{Kandinsky2.2}&& SDXL&SD2&Kandinsky2.2 \\ \hline

SDXL&\baseline{\textbf{88.99}}&\textbf{86.91}&\textbf{92.44}&\textbf{91.13}&\baseline{\textbf{73.84}}&53.36&72.61&\textbf{66.60}&\baseline{\textbf{59.16}}&\textbf{51.99}&\textbf{72.01}&\textbf{61.05}&\baseline{\textbf{77.25}}&56.53&\textbf{91.34}&\textbf{75.04}\\ 

SD2&79.68&\baseline{82.63}&85.42&82.58&65.38&\baseline{\textbf{62.64}}&66.84&64.95&40.30&\baseline{50.33}&52.66&47.76&58.46&\baseline{\textbf{75.45}}&86.67&73.53\\ 

Kandinsky2.2&58.53&50.47&\baseline{85.79}&64.93&34.33&13.67&\baseline{\textbf{79.38}}&42.46&29.76&24.98&\baseline{71.61}&42.12&14.01&16.11&\baseline{89.04}&39.72\\ \bottomrule[1.3pt]

\end{tabular}
}
\vspace{-3mm}
\caption{Evaluation of falsified regions localization performance across distinct generators. Accuracy is used for evaluation.}
\label{tab:task2_locating}
\vspace{-6mm}
\end{table*}


\emph{Intra-type performance with MLLMs.} In this task, we first evaluate the Full Synthesis type performance with VLForgery. As shown in Table~\ref{tab:task1_llava_fullsynthesis}, the Full Synthesis type comprises two subtypes: Non-existent and Id-insertion. The non-existent face subtype consists of five distinct subsets, each corresponding to a specific generator. For the non-existent subtype, training and testing within each subset surpass 99.35\% accuracy, while cross-subset testing achieves an average accuracy of 91.66\%. Conversely, cross-subtype performance varies significantly. For instance, training on the Flux subset and testing on InstantID results in a classification accuracy drop to 48.1\%.

Table~\ref{tab:task1_llava_partialsynthesis} presents the performance evaluation of the Partial Synthesis type. This type consists of three subsets, each associated with a specific generator. Training was conducted on each of the three subsets, and testing was subsequently performed across all subsets. As indicated in Table~\ref{tab:task1_llava_partialsynthesis}, training on partial synthesis samples generated by SDXL-based models yields superior generalization performance.


Based on these observations, detectors trained with full-synthesis images are typically trained based on global features, focusing on the overall structure and content of the image. Also, they primarily focus on analyzing the integrity of the entire image. However, partial synthesis images often involve modifications to small areas of the image, and these details may not be obvious in the global features. Therefore, when these detectors encounter partial synthesis images with only a small local area tampered with, they may fail to capture the subtle changes, leading to a reduced recognition rate for local tampering.

\emph{Cross-Generator Faces Classification.} We propose evaluating the generalization performance of distinct generators, encompassing both Full Synthesis and Partial Synthesis types. 
As illustrated in Table~\ref{tab:task1_llava_anydetector}, all detectors were trained on samples with partial synthesis and full synthesis from the SDXL-based model, and subsequently tested on other subsets. For each compared method, we trained for 10 epochs and validated their performance on the final epoch. VLForgery exhibits a notable advantage, especially in accurately detecting partially synthesized face images

\begin{table*}[h]
\centering
\scalebox{0.83}{
\begin{tabular}{c|ccccccccccccccc}
\toprule[1.3pt]
\multirow{1}{*}{Method} &
FS\_SDXL&FS\_SD2&FS\_SD3&FS\_IID&FS\_Kan2.2&FS\_Flux&PS\_SDXL&\_PS\_SD2&PS\_Kan2.2&Avg. \\  \hline
resnet50&99.93&99.53&84.43&98.82&99.23&99.87&11.81&6.09&76.48&75.13\\
xception&99.98&99.97&87.93&99.57&99.37&99.52&16.87&27.64&91.81&80.30\\
efficientb4&99.99&99.88&88.93&98.15&98.95&99.90&27.61&27.85&81.43&80.29\\
Guarnera \emph{et al.}&99.78&99.87&86.65&97.72&99.92&99.43&35.75&12.71&90.17&80.22\\ \hline
\baseline{VLForgery}&\baseline{93.27}&\baseline{92.47}&\baseline{93.24}&\baseline{90.42}&\baseline{89.13}&\baseline{82.43}&\baseline{63.64}&\baseline{60.20}&\baseline{73.45}&\baseline{82.03}\\ \hline
\bottomrule[1.3pt]
\end{tabular}
}
\vspace{-3mm}
\caption{Evaluation of model’s attribution accuracy on VLF.
}
\label{tab:attribution_app}
\vspace{-4mm}
\end{table*}

\begin{table*}[h]
\centering
\scalebox{0.67}{
\begin{tabular}{cccccccccccc}
\toprule[1.3pt]
 \multicolumn{3}{c|}{} & \multicolumn{9}{c}{Detection(\%)\textuparrow} \\ \cline{4-12} 
 
 \multicolumn{3}{c|}{\multirow{-2}{*}{Method}} & \multicolumn{2}{c|}{SDXL} & \multicolumn{2}{c|}{SD2} & \multicolumn{2}{c|}{Kandinsky2.2} &\multirow{2}{*}{SD3} &\multirow{2}{*}{Flus} &\multirow{2}{*}{InstantID}\\ \cmidrule(r){1-3} \cmidrule(r){4-5} \cmidrule(r){6-7} \cmidrule(r){8-9}
 
 \multicolumn{1}{c|}{Name} & VF & \multicolumn{1}{c|}{EKCot} &Partial Synthesis& \multicolumn{1}{c|}{Full Synthesis} & Partial Synthesis&\multicolumn{1}{c|}{Full Synthesis} &Partial Synthesis & \multicolumn{1}{c|}{Full Synthesis}& & &  \\ \hline
 
  \multicolumn{1}{c|}{VariantA} &&\multicolumn{1}{c|}{}&69.92&99.92&51.07&99.97&99.56&99.74&97.66&99.12&99.69\\
  
  \multicolumn{1}{c|}{VariantB} &&\multicolumn{1}{c|}{\checkmark}&68&99.99&47.64&99.98&99.18&99.99&99.98&99.98&99.84 \\ \hline
  
  \multicolumn{1}{c|}{\baseline{Ours}} &\baseline{\checkmark}&\multicolumn{1}{c|}{\baseline{\checkmark}}&\baseline{78.62}&\multicolumn{1}{c|}{\baseline{99.98}}&\baseline{66.32}&\multicolumn{1}{c|}{\baseline{99.97}}&\baseline{99.96}&\multicolumn{1}{c|}{\baseline{99.97}}&\baseline{99.98}&\baseline{99.97}&\baseline{99.57}\\ \hline
\end{tabular}
}
\vspace{-3mm}
\caption{Ablation Study of the VLForgery's forgery description generation component in the detection task. `VF', and `EkCot' represent visual input features in forgery description generation, and extrinsic knowledge-guided chain-of-thought, respectively}
\label{tab:ablation_method_detection}
\vspace{-4mm}
\end{table*}

\begin{table*}[h!]
\centering
\scalebox{0.67}{
\begin{tabular}{ccccccccccccccc}
\toprule[1.3pt]
 \multicolumn{3}{c|}{} & \multicolumn{12}{c}{Localization(\%)\textuparrow} \\ \cline{4-15} 
 \multicolumn{3}{c|}{\multirow{-2}{*}{Method}} & \multicolumn{3}{c|}{hair} & \multicolumn{3}{c|}{brows} & \multicolumn{3}{c|}{ears} & \multicolumn{3}{c}{nose} \\ \cline{1-15} 
 
 \multicolumn{1}{c|}{Name} & VF & \multicolumn{1}{c|}{EKCot} &SDXL& SD2 & \multicolumn{1}{c|}{Kandinsky2.2}&SDXL & SD2 & \multicolumn{1}{c|}{Kandinsky2.2}&SDXL & SD2\textuparrow & \multicolumn{1}{c|}{Kandinsky2.2}&SDXL & SD2 & \multicolumn{1}{c}{Kandinsky2.2} \\ \hline
 
  \multicolumn{1}{c|}{VariantA} &   & \multicolumn{1}{c|}{} &54.41& 55.06 & \multicolumn{1}{c|}{59.49} &38.63& 33.10 & \multicolumn{1}{c|}{46.76} &22.56& 18.44  & \multicolumn{1}{c|}{30.46} &38.63& 28.10 & 42.82 \\
  
  \multicolumn{1}{c|}{VariantB} &   & \multicolumn{1}{c|}{\checkmark} & 88.77 &78.51 & \multicolumn{1}{c|}{97.48} &66.09& 39.44 & \multicolumn{1}{c|}{75.61} &45.25& 34.86 & \multicolumn{1}{c|}{74.56} &40.36& 46.95 & 96.13  \\ \hline
  
  \multicolumn{1}{c|}{\baseline{Ours}} & \baseline{\checkmark} & \multicolumn{1}{c|}{\baseline{\checkmark}} & \baseline{88.99}&\baseline{86.91} & \multicolumn{1}{c|}{\baseline{92.44}} &\baseline{73.84}& \baseline{53.36} & \multicolumn{1}{c|}{\baseline{72.61}}&\baseline{59.16} &\baseline{51.99} & \multicolumn{1}{c|}{\baseline{72.01}} & \baseline{77.25}&\baseline{56.53} & \baseline{91.34} \\ \hline
\end{tabular}
}
\vspace{-3mm}
\caption{Ablation Study of the VLForgery's forgery description generation component in the localization task. `VF', and `EkCot' represent visual input features in forgery description generation, and extrinsic knowledge-guided chain-of-thought, respectively.}
\label{tab:ablation_method_localization}
\vspace{-4mm}
\end{table*}
\vspace{-1mm}
\subsubsection{Task2: Locating the Falsified Regions}
\vspace{-1mm}
We evaluate the localization capabilities of VLForgery for four distinct regions (\textit{i.e.}, nose, brow, hair, and ears), as shown in Table~\ref{tab:task2_locating}. The Partial Synthesis type contains edited faces with single-edit and multi-edit. Additionally, this type is categorized into three subtypes based on distinct generators. Each subtype was trained separately, and tested across all subtypes, with validation results tallied across the four facial regions. Note that, if a multi-edited face is detected with the result `earAndnose', but the ground truth is `earAndhair', the detection of the ear region is marked as correct, while the hair region is marked as incorrect.

\begin{figure}[h]
  \centering
  \includegraphics[width=0.47\textwidth]{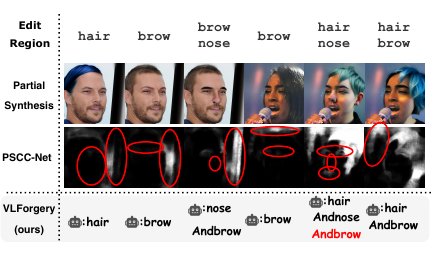}
  \vspace{-6mm}
    \caption{Comparative results of tamper facial localization capabilities between IFDL method and VLForgery. Both models were trained on partial synthesis generated by SDXL, and tested on Kandinsky2.2 samples. Mislocalized areas are marked in red.}
  \label{fig:comparison with IFDL}
  \vspace{-4mm}
\end{figure}

\begin{table}[h]
\centering
\scalebox{0.66}{
\begin{tabular}{c|cccc|cccc}
\toprule[1.3pt]
\multirow{2}{*}{Method} &
\multicolumn{4}{c|}{SD2(\%)\textuparrow}&\multicolumn{4}{c}{Kandinsky2.2(\%)\textuparrow} \\ \cmidrule(r){2-5} \cmidrule(r){6-9}

&hair&brows&ears&nose&hair&brows&ears&nose \\ \hline

PSCC-Net \cite{liu2022pscc}&53.57&39.47&41.20&29.73&60.71&68.42&63.16&44.44\\ 
\baseline{VLForgery}&\baseline{\textbf{86.91}}&\baseline{\textbf{53.36}}&\baseline{\textbf{51.99}}&\baseline{\textbf{56.53}}&\baseline{\textbf{92.44}}&\baseline{\textbf{72.61}}&\baseline{\textbf{72.01}}&\baseline{\textbf{91.34}}\\ \bottomrule[1.3pt]

\end{tabular}
}
\vspace{-3mm}
\caption{\centering{Evaluation of falsified regions localization performance across distinct generators with VLForgery and IFDL method.}}
\label{tab: quantity comparison with IFDL}
\vspace{-6mm}
\end{table}

\emph{Why Use Natural Language Instead of Masks for Localization Results?}
Fig.~\ref{fig:comparison with IFDL} illustrated a comparison of localization results between VLForgery and PSCC-Net (IFDL method). For subtle falsified facial regions, such as brows and nose, IFDL methods often struggle to identify them accurately, being easily influenced by extraneous features from the image background, which leads to incorrect localization. In contrast, for facial localization, where tampered areas are limited, VLForgey leverages natural language descriptions instead of mask images to reduce the difficulty of localization. This approach significantly enhances localization accuracy and minimizes susceptibility to interference from background features. Furthermore, to quantify the performance gap in localization between ours and the IFDL method, we selected 1,000 samples from the Partial Synthesis test set. We used the IFDL method to generate mask images, which were then manually evaluated. A localization was deemed correct if any part of the manipulated area was slightly visible in the mask image; otherwise, it was considered incorrect. As shown in Table~\ref{tab: quantity comparison with IFDL}, our method demonstrates a clear advantage in accurately localizing manipulated regions.

\emph{Cross-Generator Falsified Region Localization.} As observed in Table~\ref{tab:task2_locating}, localization accuracy for hair consistently remains the highest, followed by the nose, brows, and then the ears. For instance, when trained and tested on SDXL samples, the hair localization accuracy reached 88.99\%. In contrast, localization accuracy for smaller regions, such as the ears, decreases to 59.16\%. 


\vspace{-1mm}
\subsubsection{Task3:Attributing Source Models of Synthesis Faces}
\vspace{-1mm}
In this task, we focus on evaluating the VLForgery's capability to capture the synthesis semantic patterns of different generators.  We fine-tuned VLForgery on the VLF training set, which includes all subsets, and tested its performance across each generator for the two synthesis types. As shown in Tab.~\ref{tab:attribution_app}, the attribution performance for full synthesis is superior, with an average accuracy of 91.71\%. However, attribution accuracy for partial synthesis falls below 70\%. We attribute this disparity to the detection difficulty of partial synthesis.

\vspace{-1mm}
\subsection{Ablation Study}
\vspace{-1mm}
Two main factors affect VLForgery forensics performance: (1) the impact of EkCot compared to outputting only forgery detection results, and (2) the influence of additional visual input features on forensics performance in EkCot. We conducted ablation studies for both detection and localization tasks. Table~\ref{tab:ablation_method_detection} presents the ablation results for the detection task, demonstrating a significant performance improvement in detecting partially synthesized images when using the EkCot. Table~\ref{tab:ablation_method_localization} displays the ablation results for the localization task, where the model's localization capability significantly decreased without EkCot. And compared to the forgery description generation method without visual assistance, the model performed better on SDXL and SD2 samples, though it showed some limitations on Kandinsky 2.2 generated samples.
\vspace{-1mm}
\subsection{Qualitative Study}
\vspace{-1mm}
The primary advantage of the multimodal large language models (MLLMs) is their flexible output, allowing them to provide a comprehensive analysis of the synthesized images. As shown in Fig.~\ref{fig:comparison with advanced LLMs}, we compared VLForgery with some existing MLLMs, demonstrating that VLForgery enables detailed capabilities, including forgery analysis, localization of falsified regions, and synthesis attribution. 


\vspace{-2mm}
\section{Conclusion}
\vspace{-2mm}
In this paper, we introduce VLForgery, a framework designed for AI-generated faces in fine-grained forensics scenarios. To address the lack of partially synthesized face datasets, we constructed VLF. Additionally, we introduce EkCot, which provides a fine-grained analysis of forgery artifacts. The satisfying performance of the proposed framework is justified by extensive evaluations.
